# Dynamic imaging and characterization of volatile aerosols in e-cigarette emissions using deep learning-based holographic microscopy

Yi Luo[1,2,3,+]

Yichen Wu [1,2,3,+]

Liqiao Li [4,+]

Yuening Guo[4]

Ege Çetintaş[1,2,3]

Yifang Zhu[4]

Aydogan Ozcan[1,2,3,5,*]

[1]Electrical and Computer Engineering Department, University of California, Los Angeles, California 90095, USA

[2]Bioengineering Department, University of California, Los Angeles, California 90095, USA

[3]California Nano Systems Institute (CNSI), University of California, Los Angeles, California 90095, USA

[4]Department of Environmental Health Sciences, University of California, Los Angeles, California 90095, USA

[5]David Geffen School of Medicine, University of California, Los Angeles, California 90095, USA

[+]Equal contribution authors

[*]Correspondence: Prof. Aydogan Ozcan

E-mail: ozcan@ucla.edu

Address: 420 Westwood Plaza, Engr. IV 68-119, UCLA, Los Angeles, CA 90095, USA

Tel: +1(310)825-0915

Fax: +1(310)206-4685

# Abstract

Various volatile aerosols have been associated with adverse health effects; however, characterization of these aerosols is challenging due to their dynamic nature. Here we present a method that directly measures the volatility of particulate matter (PM) using computational microscopy and deep learning. This method was applied to aerosols generated by electronic cigarettes (e-cigs), which vaporize a liquid mixture (e-liquid) that mainly consists of propylene glycol (PG), vegetable glycerin (VG), nicotine, and flavoring compounds. E-cig generated aerosols were recorded by a field-portable computational microscope, using an impaction-based air sampler. A lensless digital holographic microscope inside this mobile device continuously records the inline holograms of the collected particles. A deep learning-based algorithm is used to automatically reconstruct the microscopic images of e-cig generated particles from their holograms, and rapidly quantify their volatility. To evaluate the effects of e-liquid composition on aerosol dynamics, we measured the volatility of the particles generated by flavorless, nicotine-free e-liquids with various PG/VG volumetric ratios, revealing a negative correlation between the particles' volatility and the volumetric ratio of VG in the e-liquid. For a given PG/VG composition, the addition of nicotine dominated the evaporation dynamics of the e-cig aerosol and the aforementioned negative correlation was no longer observed. We also revealed that flavoring additives in e-liquids significantly decrease the volatility of e-cig aerosol. The presented holographic volatility measurement technique and the associated mobile device might provide new insights on the volatility of e-cig generated particles and can be applied to characterize various volatile PM.

# Introduction

Exposure to particulate matter (PM) has been associated with various adverse health effects in epidemiological studies[1,2]; some of these particles contain a large fraction of volatile or semi-volatile materials, such as emissions generated by cooking[3], vehicles[4], and usage of tobacco products[5,6]. Since the particle dynamics-related information derived from non-volatile aerosols cannot be applied to volatile or semi-volatile emission sources, there is a need to better understand and measure PM volatility and related dynamic behaviour for assessing their exposure and potential health impact.

The dynamic evaporation of a volatile particle raised enduring research interest that can be traced back to the 19$^{th}$ century[7]. Thermal denuders were used as a potential candidate, providing inline saturation pressure measurements, especially in aerosol studies[4,8,9]. A thermal denuder compares particle size distribution and concentration of a cloud of volatile particles before and after they evaporate under precise control of the heating temperature and time. Thermal dynamic models are used to map differences in particle number and size distribution to the physical parameters of volatile particles. Unfortunately, such models are still missing for aerosols that have complex chemical compositions[10]. Another way of measuring a particle's volatility is to image its geometrical/structural change when it is sitting on a transparent substrate[7,11]. However, these earlier methods are used to measure relatively large particles that have a footprint on the order of square-millimeters, and suffer from extremely low throughput.

Here we report a high-throughput volatile particle measurement system that is based on computational inline holography and deep learning (Fig. 1), which was applied for dynamic imaging and characterization of volatile aerosols created by electronic cigarettes (e-cigs). An impactor-based portable air quality monitoring device[12] was used to sample aerosols onto a transparent sticky sampling pad, at a throughput of 13 L/min. Time-lapsed inline holograms of the sampled particles (typically more than ten thousand particles per experiment) were recorded at 2 frames per second (fps) across a field-of-view of 4 mm$^2$. A deep neural network[13] was used to reconstruct the acquired inline holograms, revealing the phase and amplitude image of each particle as a function of time, which was further processed to measure its volatility. This system was applied to characterize e-cig generated aerosols, which are highly dynamic[14–16] due to the volatile materials used in e-cig liquid mixtures (e-liquids).

E-cigs have gained worldwide attention, primarily due to their unprecedented popularity among never-smoking adolescents and young adults over the last decade[17–20]. The use of an e-cig (or *vaping*) generates an inhalable aerosol by heating and vaporizing an e-liquid, which typically uses propylene glycol (PG) and vegetable glycerin (VG) as the solvents to dilute nicotine and flavoring compounds. Many previous studies have reported the potential adverse health effects and toxicity of e-cig aerosols [21–26]. The recent outbreak of e-cig or vaping product use-associated lung injury (EVALI) has particularly raised public health concerns[27–29]. Previously, Li *et al.*[30] reported that the PG/VG ratio and the addition of nicotine in the e-liquid changed the particle loss rates based on the slope of the log-normalized PM concentrations, suggesting that the evaporative properties may be linked to the constituents of the e-liquid. However, this traditional method was relatively time-consuming and indirect since the particle loss rate estimation could not directly provide the volatility information.

In this work, we used a holographic mobile imaging device to measure the volatility of aerosols generated by different e-liquid compositions using a customized puffing machine to validate the effects of e-liquid composition on aerosol volatility at the micro-scale. Our experimental results revealed a negative correlation between the e-cig aerosol's volatility and the volumetric ratio of VG in nicotine-free e-liquids. The addition of other chemical, e.g., nicotine and flavoring chemicals, introduced significant changes in the observed volatility patterns. Nicotine was found to dominate the evaporation dynamics of e-cig emission and overwhelm the impact of VG volumetric ratio. The addition of flavoring chemicals into the e-liquids also significantly decreased the volatility of e-cig generated aerosols. The presented holographic volatility measurement technique with its mobile interface provides direct and high-throughput quantification of the volatility of e-cig generated aerosols, and can be broadly applied to rapidly characterize various volatile PM.

## Results and Discussion

**Real-time imaging and quantification of e-cig generated particles for different e-liquid compositions**

To investigate the impact of e-liquid composition on e-cig generated aerosol volatility, we created aerosols using a customized puffing machine mimicking human vaping (see Figs. 1(a-b) and the Methods section for details). An Aerodynamic Particle Sizer (APS) and a Condensation Particle Counter (CPC) were used to provide independent measurements of the particle size distribution and the particle number concentration (PNC), respectively. The e-cig generated aerosol was also imaged with a holographic PM monitoring device (termed c-Air) which was used to quantify its volatility, particle by particle. In c-Air, aerosols were collected using an impactor operating under a flow rate of 13 L/min. Particles captured by the impactor land on top of a transparent sticky pad, and time-lapsed inline holograms of the captured particles are recorded at 2 fps. After the air sampling is complete (~180 s), an additional 360 holographic frames were acquired at the same frame rate, to sample the evaporation process of the captured volatile particles. A deep neural network[13] was used to perform auto-focusing and phase recovery to generate the phase and amplitude images of all the collected particles as a function of time (see Methods section for details). The acquired holographic image sequence of every detected particle (e.g., Fig 1(d)) was then cropped out to determine the volatility per particle, covering a sample field-of-view of 4 mm$^2$. The volatility of a sessile droplet landing on a substrate can be described by the decay rate of the contact angle ($\theta$) between its edge and the substrate[11,31]. Therefore, using these holographic images, the change in the contact angle of each particle with respect to the substrate was extracted as a function of time (*t*) using the reconstructed phase channel (see Methods section for details). The contact angle decay rate $K_i$ (rad/s) of each detected particle *i* was fitted using a linear model, i.e.,

$$\theta_i(t) = \theta_{i0} - K_i \cdot t \qquad (1)$$

For each e-liquid sample, three independent measurements were conducted and *n* different particles (typically *n* > 10,000 per measurement) were collected and analyzed. Measured angle decay rates for each particle ($K_1 \ldots K_n$) were fit to a Gaussian distribution, and the mean value of the fitted Gaussian distribution was treated as the calculated volatility constant *K* across all the measured particles.

Based on this analysis, PM volatility statistics and size distribution measurements for homemade and commercial e-liquids with varying PG/VG ratios are reported in Fig. 2 and Supplementary Table S1. As the PG/VG ratio of homemade (or commercial) e-liquid decreased from 70/30 to 0/100, the measured volatility constant *K* decreased from 0.058±0.0002 rad/s (0.050±0.002 rad/s) to 0.038±0.0023 rad/s (0.043±0.001 rad/s). For the homemade e-liquids, we found that the volatility constant is negatively correlated with the volumetric ratio of VG within the e-liquid ($p<0.05$) except for the case of PG/VG = 100/0, suggesting that e-cig generated particles become less volatile as the proportion of VG increases in the e-liquid. This might be due to the fact that VG has substantially lower vapor pressure than PG (i.e., $P_{sat\ (VG)} \approx$ 20 Pa while $P_{sat\ (PG)} \approx 0.1$ Pa at standard conditions), and this may lower the volatility of e-cig aerosols as the PG/VG ratio is reduced, in accordance with the Raoult's law.

It is also noteworthy to mention that the volatility of aerosols produced from pure PG may be considerably underestimated in our measurements at 2 fps because of its fast-evaporating nature. As required by the sampling protocol of this study, the generated particles underwent a mixing and dilution process in the test chamber for ~26-30 s before they are sampled by c-Air (see Methods section). As a result, the pure-PG particles that were captured and processed by c-Air were substantially lower in count. The PNC generated from pure PG was 1 to 2 orders of magnitude lower than other combinations of PG and VG, under the same particle generation protocol, i.e., a single puff of the e-cig; see Figure 2 (a). In accordance with our observations, PG generated aerosols were previously reported to have higher particle loss rates due to their fast evaporating nature compared to other e-liquid compositions[30].

c-Air volatility measurements can also be compared against the PNC and particle size distributions that were simultaneously recorded for each sample (see Figs. 2(a-b)). As the PG/VG ratio shifted from 100/0 to 50/50, PNC inside the test chamber increased by more than 2 order of magnitudes (e.g., for homemade liquids from 435±124 to 1.09±0.09×$10^5$ #/cm$^3$). Further increase of VG concentration did not lead to a major PNC increase. The rapid PNC increase at lower VG levels may indicate that mostly the VG content in the e-liquid contributed to the aerosols that were generated. The particle size distributions in Figure 2(b) report the particle size (in μm) in x-axis and the normalized PNC (dN/dLogDp) in y-axis. As the PG/VG ratio in the e-liquid decreased, the mode of particle size distribution shifted from < 0.5 μm to ~ 1.0 μm. The pure-PG particles in the e-cig vapor with their smaller size likely have higher volatility, as smaller liquid particles with greater vapor pressure evaporate faster due to the Kelvin effect[32].

Overall, our experimental results and analyses indicate that e-cig generated particles evaporate slower with decreasing PG/VG ratios, along with a reduction in volatility with increasing particle size.

**Impact of nicotine on the evaporation behavior of e-cig generated aerosols**

While the solvents PG and VG are used to provide a tobacco-like smoking experience, e-cigs are also used as electronic nicotine delivery devices with varying levels of nicotine[33]. The effect of nicotine level on particle volatility is a key aspect in understanding e-cig aerosol dynamics. We examined the impact of three different nicotine levels (i.e., 0%, 1.2% and 2.4% in mass concentration) on the volatility of e-cig aerosols generated by five different combinations of PG/VG ratios, as shown in Figure 3. The addition of nicotine significantly reduced the volatility of particles with high PG composition ($p<0.05$). The measured volatility for pure PG particles (and 70/30 PG/VG particles) reduced from $0.036 \pm 0.019$ rad/s ($0.058 \pm 0.001$ rad/s) to $0.022 \pm 0.002$ rad/s ($0.043 \pm 0.003$ rad/s), when 1.2% of nicotine was added into a nicotine-free e-liquid. The volatility across different PG/VG ratios remained at a similar level for 1.2%-nicotine added e-liquids with a PG/VG ratio that is smaller than 70/30. Moreover, increasing the nicotine concentration from 1.2% to 2.4% did not further reduce the volatility for different PG/VG compositions.

These observations are also supported by the PNC and particle size distribution measurements (Supplementary Fig. S2). For pure-PG particles, adding 1.2% nicotine increased the PNC by one order of magnitude, from $4.35\times10^2\pm1.24\times10^2$ to $4.29\times10^3\pm6.63\times10^2$ #/cm$^3$. Furthermore, adding 1.2% nicotine increased the PNC produced by 70/30 PG/VG e-liquid to the same order of magnitude as those with higher VG compositions generated. Similar to c-Air measurement results, no statistically significant change in PNC was observed between 1.2% and 2.4% nicotine-added e-liquid samples. Interestingly, the decrease in the volatility that we observed in PG/VG mixtures with increasing VG levels was no longer evident after the nicotine was introduced into the e-liquid. In a simple two-component system, the PG/VG ratio determined the saturation vapor pressure of the e-liquid mixture, which dominated the volatility behavior of e-cig generated aerosols. However, the addition of nicotine fundamentally alters the evaporation process, forming a more complex mixture with a different volatility behavior as we observed.

**Impact of flavoring additives on the volatility of e-cig generated particles**

In addition to nicotine levels, we also evaluated how flavoring additives changed the evaporation dynamics of e-cig generated particles. Both commercially-available flavored e-liquids (e.g., tobacco, menthol, and strawberry flavors) and flavorless (no flavoring additives added) e-liquids were tested, with the flavorless e-liquid serving as the baseline for volatility comparisons. To avoid the confounding factors from different PG/VG ratios and nicotine levels, we used 50/50 PG/VG nicotine-free e-liquids for all four tests. Ours results, summarized in Fig. 4, indicate that flavoring additives significantly changed the volatility of e-cig generated aerosols. The volatility constant of flavorless e-liquid generated aerosols was measured to be $0.044\pm0.0002$ rad/s, while the volatility constants of the three flavored e-liquids were measured as $0.043\pm0.002$ rad/s for tobacco flavor, 0.041±0.002 rad/s for menthol flavor, and 0.041±0.0004 rad/s for strawberry flavor. A paired t-test was performed to compare the distribution of volatility constants between the flavorless e-liquid and each one of the flavored e-liquids, respectively, where a statistically significant change was obtained for all the flavors (each with $p<0.05$). Overall, these measurements reveal that the addition of these flavoring compounds into the e-liquid reduced the volatility of the e-cig generated aerosols.

Although a statistically significant change in volatility was observed in our c-Air measurements, the impact of e-liquid flavoring on the particle size distribution and PNC measurements were minimal (Figure 4(b)). When compared with the strong shifts observed in the particle size distributions due to the varying PG/VG ratios shown in Figure 2, the subtle changes introduced by flavoring additives indicate that the particle generation process is mainly governed by the PG/VG ratio (in the absence of nicotine). Nevertheless, the unknown and complicated chemical compositions and concentrations of these flavoring additives still complicate the evaporation behavior of e-cig generated particles, as shown in our volatility measurements.

**Future work and conclusions**

In this work, we studied the evaporation dynamics of PM generated by e-cigs and we believe that understanding of these dynamics might open up new avenues for e-cig related research. Benefiting from its dynamic measurement capability, c-Air platform can also be used in the studies that examine second-hand vaping aerosols, where the aerosol volatility could heavily affect their behavior in the environment, resulting in different exposure mechanisms for public[14,34]. The presented method and measurement device can also serve as a powerful tool in investigating particle/gas partitioning of any atmospheric aerosols, in addition to e-cig generated aerosols. Therefore, the methodology used in this work can be applied to various fields that require dynamic measurements of volatile particles using portable devices.

In conclusion, we examined the volatility of e-cig generated aerosols using lensless microscopy and deep learning. A negative correlation between e-cig generated particle volatility and VG concentration in the e-liquid was revealed. The addition of other chemicals (e.g., nicotine and flavoring compounds) reduced the overall volatility of e-cig generated aerosols. The results obtained with the high-throughput c-Air device are consistent with previous studies that used traditional measurements methods. The presented approach can help us better examine the dynamic behaviour of e-cig aerosols in a high-throughput manner, potentially providing important information for e-cig exposure assessment via e.g., second-hand vaping.

## Materials and Methods

**E-cigarette sample preparation**

In this study, both homemade and commercially-available e-liquids with five different PG/VG volumetric ratios (100/0, 70/30, 50/50, 30/70 and 0/100) and with 0.0%, 1.2% and 2.4% nicotine (mass concentration) were tested. Homemade e-liquids were prepared from individual chemicals, including PG ($C_3H_8O_2$, ≥99.5%, Sigma-Aldrich), VG ($C_3H_8O_3$, ≥ 99.5%, Sigma-Aldrich) and nicotine ($C_{10}H_{14}N_2$, ≥99%, Sigma-Aldrich). To ensure the quality of the e-liquids, all the studied e-liquids were well-mixed and prepared within 7 days before each experiment. All commercial e-liquids used in this study were purchased from VaporVapes Inc. (Sand City, CA).

**Puffing machine**

The puffing machine was composed of an acrylic holder, an e-cig device with a refillable tank, a power source, and an Arduino UNO R3 microcontroller. The e-cig device used in the study was the Vapor-fi Volt II Hybrid Tank[30]. It was selected as a representative tank-type e-cig device on the market[35,36]. The e-cig tank was equipped with a 0.5-Ω heating coil and powered at 15 W (i.e., 6 V and 2.5 A).

**Aerosol characterization**

The test chamber used in this study has a volume of 0.46 m$^3$. The air exchange rate of the chamber was maintained at 1 h$^{-1}$ (i.e., the air inside the chamber gets replaced once every hour). The dilution ratio of the chamber is approximately 6900:1[30]. Temperature and humidity were measured by an indoor air quality monitor (Q-Trak 7575, TSI Inc.) inside the test chamber. A portable Condensation Particle Counter (CPC 3007, TSI Inc.) was used to measure PNC. To ensure data quality, the background PNC inside the chamber was kept at <100 particles/cm$^3$ before each sampling session. An Aerodynamic Particle Sizer (APS 3321, TSI Inc.) was used to measure the particle size distribution ranging from 0.5 μm to 19.8 μm.

**Time-lapsed imaging of e-cig generated aerosols**

A portable computational air quality monitor (c-Air) was developed to track the e-cig generated aerosols based on lensless digital holography[37,38] (Fig. 1). The c-Air device was connected to the e-cig test chamber through a 1/4-inch inner diameter Tygon tube. A miniaturized vacuum pump (M00198, GTEK Automation) inside c-Air sampled aerosols from the tubing through a disposable impactor at 13 L/min, where the aerosols within the air stream landed on a sticky transparent coverslip. The c-Air device used a vertical-cavity surface-emitting laser (VCSEL) diode (OPV300, TT Electronics, peak wavelength $\lambda = 850\ nm$), which illuminated the coverslip from above and created inline holograms of the collected aerosols. These digital holograms were recorded as time-lapsed images at 2 fps by a complementary metal–oxide–semiconductor (CMOS) image sensor chip (Sony IMX219PQ, pixel pitch 1.12 µm) below the impactor, while the aerosols were being sampled. The c-Air device was powered by a Lithium polymer (Li-po) battery (Turnigy Nano-tech 1000mAh 4S 45~90C Li-po pack) and controlled by a Raspberry Pi Zero W single board computer.

**Hologram reconstruction**

The CMOS image sensor that was used to capture the holograms of e-cig generated aerosols was a Bayer color image sensor, which has slightly different responses for the four Bayer channels under the infrared illumination (850 nm). To correct for this difference, we performed a wavelet transform (using 5-th order symlet) on each of the Bayer channels individually to estimate a low frequency back-ground shade image. Each channel was divided by their corresponding shade image to correct for the illumination shade and channel imbalances due to the Bayer filters. These corrected holograms were each digitally back-propagated by 750 µm to roughly reach the top of the coverslip surface using the angular spectrum method[39,40]. Zero padding was used in the angular spectrum domain to effectively up-sample the back-propagated image by two times in the spatial domain. The twin image noise and defocus artifacts in the back-propagated holograms were removed by a convolutional neural network (CNN)[13] that was specifically trained on e-cig generated aerosols to transform a randomly defocused, back-propagated hologram image to a phase recovered, in-focus image, which achieves holographic phase recovery and autofocusing at the same time and generates an extended depth of field image reconstruction. The network structure was the same as in reference [13]. To provide training data for this CNN, we used experimental data for the e-cig samples that were captured independent of the test results reported in this manuscript. Each hologram was split into smaller patches of 512×512 pixels.

These patches were each auto-focused and phase recovered using a modified Gerchberg-Saxton error reduction-based phase recovery algorithm[41–43] which was only used during the CNN training phase. The original hologram patches were randomly back-propagated to an axial range of ~200 μm above and below the determined focal plane, which were used as input to the network during its training. The network was built on TensorFlow. The training took ~30 hours for ~ 100 epochs on a laptop computer with a 3.60 GHz CPU, 16GB RAM, and an Nvidia GeForce GTX 1080Ti GPU. After the training phase, which is a one-time effort, the trained CNN was blindly used to auto-focus and phase-recover the images of the captured particulate matter. The network inference time is < 0.1 s for a back-propagated hologram patch of 512×512 pixels, and ~10 s for a full-FOV image of 3280×2464 pixels.

**Aerosol detection**

A threshold level that equals to five standard deviations away from the image mean was applied to the real and imaginary channels of each reconstructed image and a spatial mask was generated as the union of the thresholded real and imaginary channels to mark the detected particle locations. Spatial masks for each time frame were stacked together, forming a space-time mask stack. Then, individual particles were detected by finding continuous traces within the space-time mask stacks. Empirically, we have seen that a volatile particle evaporates typically within 5-30 frames (2 - 15s), as shown in e.g., Supplementary Movie S1, whereas a non-volatile particle remains unchanged, in terms of volume and shape, during the entire imaging process. As a result, cropped image sequences, which are found using the space-time masks, were used for further analysis of individual aerosol's volatility, as detailed below.

**Geometrical parameter fitting and extraction**

The collected volatile particles were assumed to be spherical caps positioned on the impactor surface. Particle volume ($V_i$) was defined as an optical phase integral inside the identified mask area of a given image frame $\Omega_i$, and was calculated using a refractive index difference of $\Delta n = 0.4$ with respect to air, which is a typical value for PG and VG[44,45], i.e.

$$V_i = \frac{\lambda}{2\pi\Delta n} \cdot \int_{\Omega_i} ang(s)\, ds \tag{2}$$

The maximum height $h_i$ for each particle was calculated using the maximum phase value inside the mask, i.e.

$$h_i = \frac{\lambda}{2\pi\Delta n} \cdot \max(ang(s)) \tag{3}$$

The contact angle for each individual particle (defining a spherical cap with $h_i$ maximum height on the substrate) was calculated as:

$$\theta_i = \arccos\left(1 - \frac{3h_i^2}{\frac{3V_i}{\pi h_i} + h_i^2}\right) \tag{4}$$

A linear model was used to fit the contact angle across different frames and the contact angle decay rate, $K_i$, was extracted to quantify the particle volatility, as detailed in Eq. (1).

# List of figures

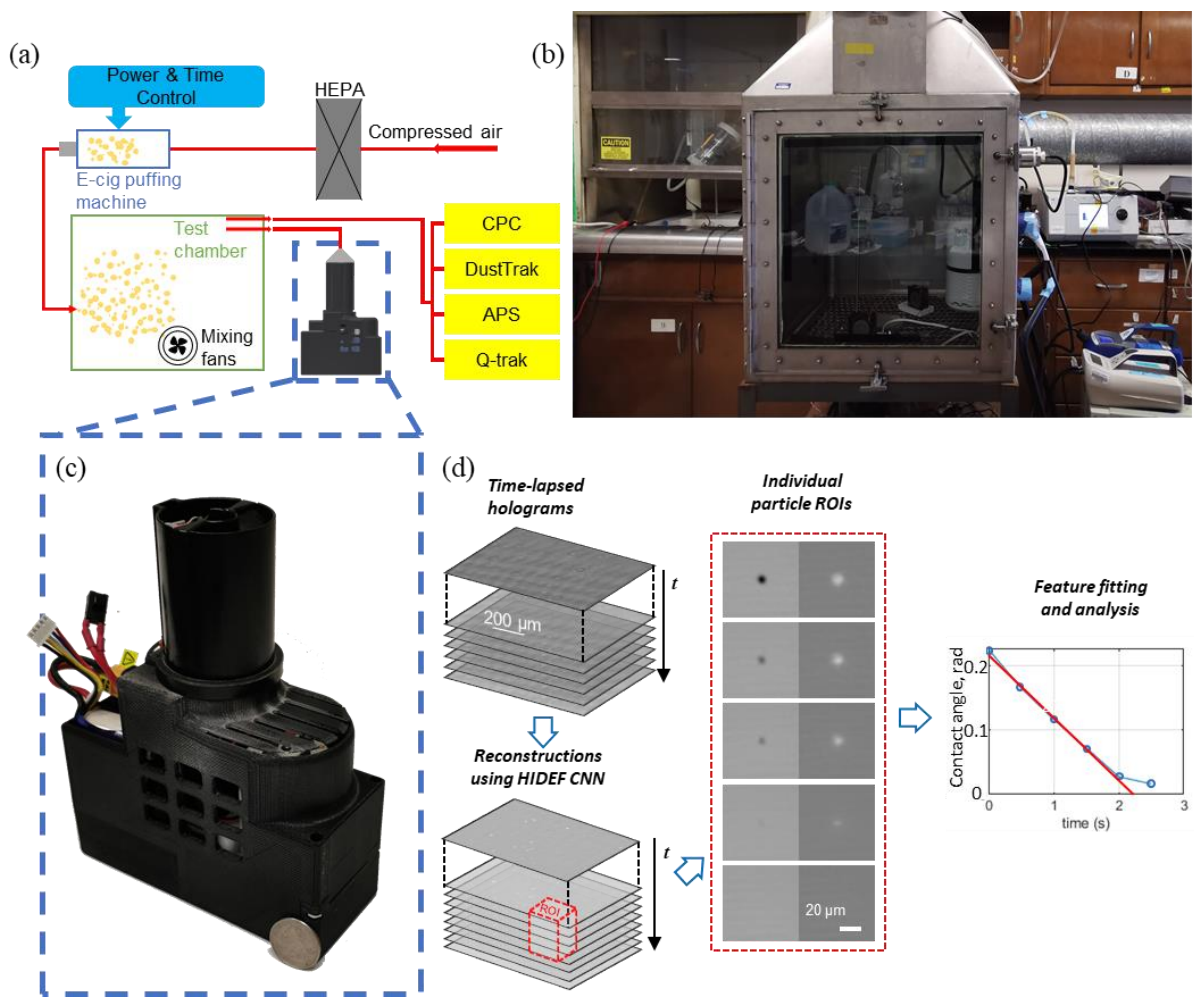

**Figure 1. Computational imaging system for directly measuring the volatility of e-cig generated aerosols.** A customized puffing machine was used to simulate human puffing of e-cig. Puffing generated e-cig aerosols were sampled by lens-free computational microscopy and the volatility of each captured particle was quantified. **a.** Schematic drawing of the puffing machine and the computational imaging system. **b.** A photograph of the presented system. **c.** A photograph of the computational imaging system, c-Air. **d.** Image processing pipeline for measuring a dynamic particle's volatility.

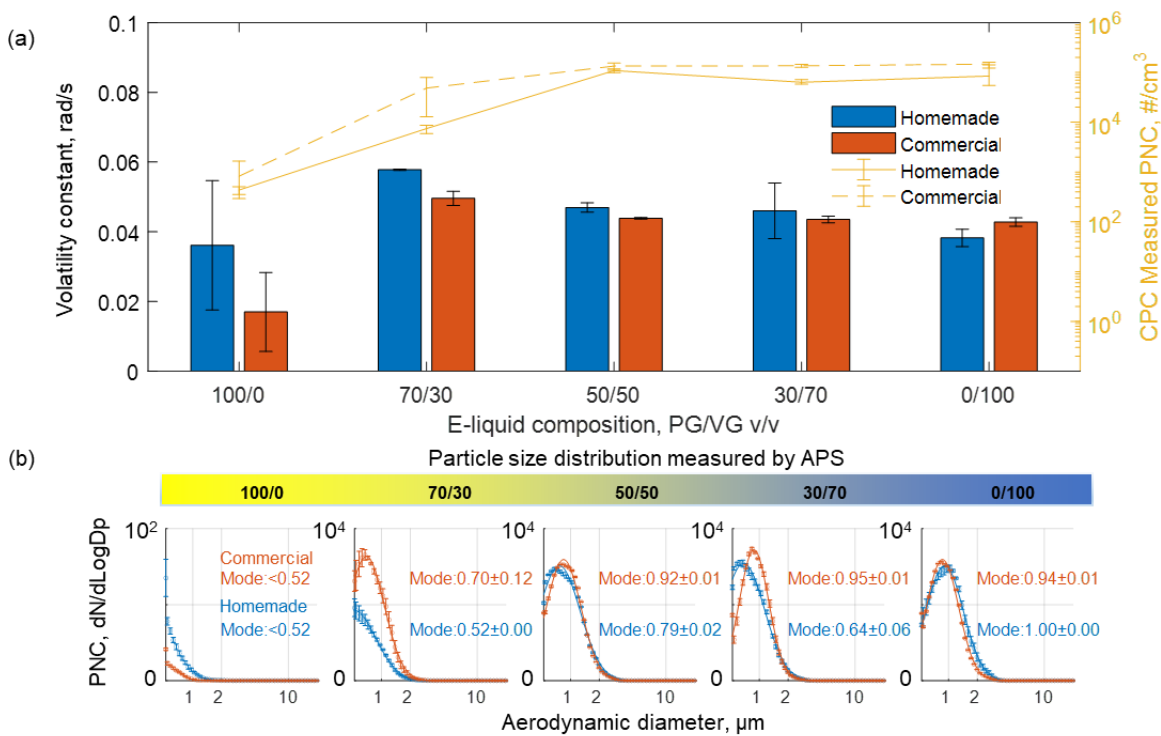

**Figure 2. Volatility measurements and particle size distribution of particles generated by nicotine-free e-liquids with different PG/VG compositions (PG/VG ratio = 100/0, 70/30, 50/50, 30/70 and 0/100). a.** Volatility measurement results of different e-liquids by c-Air and particle number concentration measurements by CPC. **b.** Particle size distributions measured by APS. A log-normal distribution was fitted to each particle size distribution to calculate the mode diameter. The histograms of the measured volatility constants and size distributions can be found in Supplementary Figure S1.

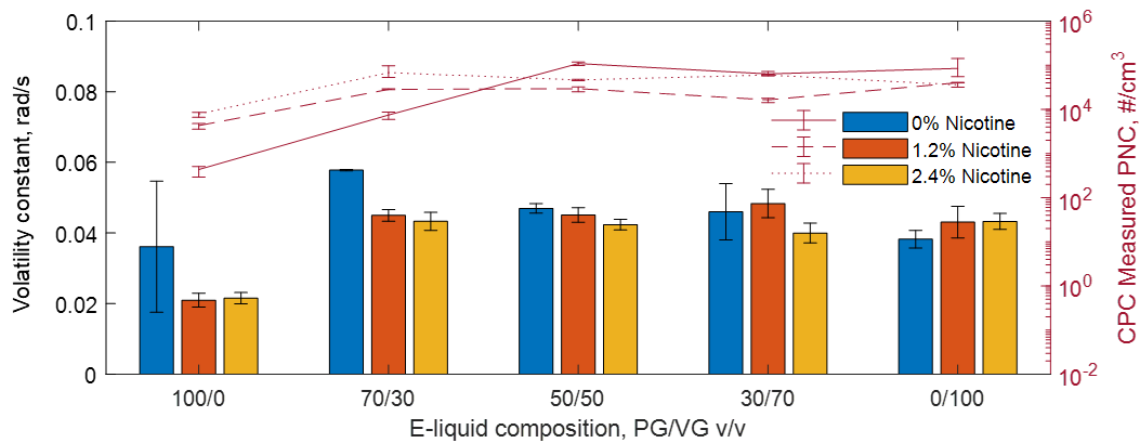

**Figure 3. Volatility measurements of particles generated by puffing homemade e-liquids with different PG/VG compositions (PG/VG ratio = 100/0, 70/30, 50/50, 30/70 and 0/100) and nicotine levels (0%, 1.2% and 2.4%).** Volatility histograms are shown in Supplementary Figure S2.

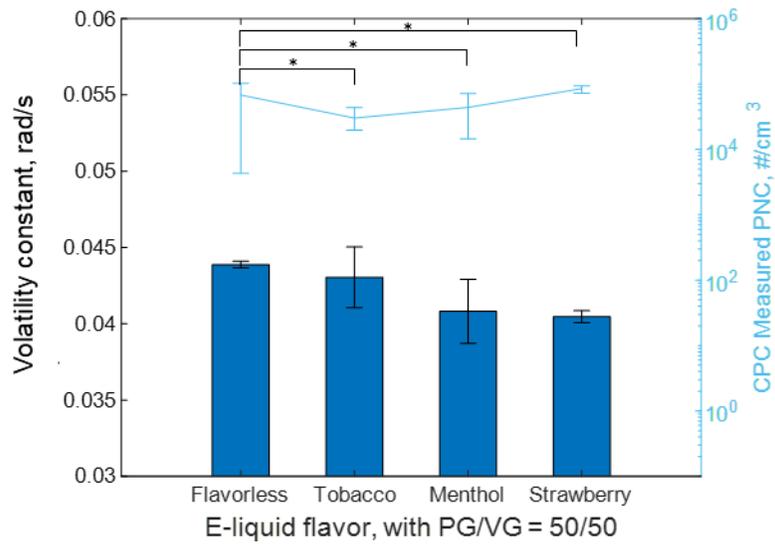

**Figure 4. Volatility measurements of particles generated by puffing nicotine-free commercial e-liquids with different flavors (flavorless, tobacco, menthol, and strawberry flavor, with PG/VG = 50/50).** Paired t-test *p<0.05. Volatility histograms are shown in Supplementary Figure S3.